\documentclass[10pt,twocolumn,letterpaper]{article}

\usepackage{cvpr}              

\usepackage{multirow}

\definecolor{cvprblue}{rgb}{0.21,0.49,0.74}
\usepackage[pagebackref,breaklinks,colorlinks,allcolors=cvprblue]{hyperref}
\usepackage[margin=1in]{geometry}
\usepackage{graphicx}
\usepackage[most]{tcolorbox} 
\usepackage{xcolor}
\usepackage{lmodern}         
\usepackage{caption}

\tcbset{
  panelA/.style={
    enhanced,
    colback=blue!5, colframe=blue!60!black,
    boxrule=0.6pt, arc=2mm, left=2mm, right=2mm, top=1mm, bottom=1mm,
    fonttitle=\bfseries, title={Vanilla SFT},
    drop shadow
  },
  panelB/.style={
    enhanced,
    colback=orange!10, colframe=orange!70!black,
    boxrule=0.6pt, arc=2mm, left=2mm, right=2mm, top=1mm, bottom=1mm,
    fonttitle=\bfseries, title={SFT-SWL},
    drop shadow
  },
  imageFrame/.style={
    enhanced,
    colback=gray!3, colframe=gray!55,
    boxrule=0.6pt, arc=3mm, left=2mm, right=2mm, top=2mm, bottom=2mm,
    drop shadow
  },
  qaBubble/.style={
    enhanced,
    colback=white, colframe=gray!55,
    boxrule=0.6pt, arc=2mm, left=2mm, right=2mm, top=1mm, bottom=1mm,
  },
}

\def\paperID{11360} 
\def\confName{CVPR}
\def\confYear{2026}

\title{Balanced Thinking: Improving Chain of Thought Training in Vision Language Models}

\author{Shaked Perek\textsuperscript{*} \and
Ben Wiesel\textsuperscript{*} 
\and
Avihu Dekel \and
Nimrod Shabtay \and
Eli Schwartz \\
IBM Research \\
{\tt\small shaked.perek@il.ibm.com} \\
{\small \textsuperscript{*}Equal contribution}
}

\begin{document}
\maketitle

\begingroup
\renewcommand{\thefootnote}{}
\footnotetext{Code: \url{https://github.com/shakedpe/scale}}
\endgroup

\begin{abstract}

Multimodal reasoning in vision-language models (VLMs) typically relies on a two-stage process: supervised fine-tuning (SFT) and reinforcement learning (RL). In standard SFT, all tokens contribute equally to the loss, even though reasoning data are inherently token-imbalanced. Long \texttt{<think>} traces overshadow short but task-critical \texttt{<answer>} segments, leading to verbose reasoning and inaccurate answers.
We propose a \textbf{SCALe – Scheduled Curriculum Adaptive Loss} that explicitly separates supervision over reasoning and answer segments using dynamic, length-independent weighting. 
Unlike vanilla SFT, which overweights the \texttt{<think>} segment, SCALe-SFT gradually shifts the  focus from \texttt{<think>} to \texttt{<answer>} throughout training
via a cosine scheduling policy, encouraging concise and well-grounded reasoning. 
We evaluate SCALe across diverse benchmarks and architectures. Results show that SCALe consistently improves accuracy over vanilla SFT and matches the performance of the full two-phase SFT + GRPO pipeline while requiring only about one-seventh of the training time, making it a lightweight yet effective alternative.
When combined with GRPO, SCALe achieves the best overall performance, highlighting its value both as a standalone method and as a strong foundation for reinforcement refinement.



\end{abstract}    
\section{Introduction}
\label{sec:intro}

Modern vision-language models (VLMs) are evolving from direct answer generators into reasoning agents capable of structured reasoning, producing intermediate chains of thought (CoT) that connect visual evidence with textual context to derive final answers \cite{lin2025mind, guo2025deepseek, plaat2024reasoning}. Effective reasoning enables these models to identify relevant visual cues, form coherent hypotheses, and reach consistent conclusions, which are essential for accuracy, interpretability, and reliability in real-world applications \cite{wang2024measuring}.

\begin{figure}[t]  
  \centering
  \includegraphics[width=\linewidth]{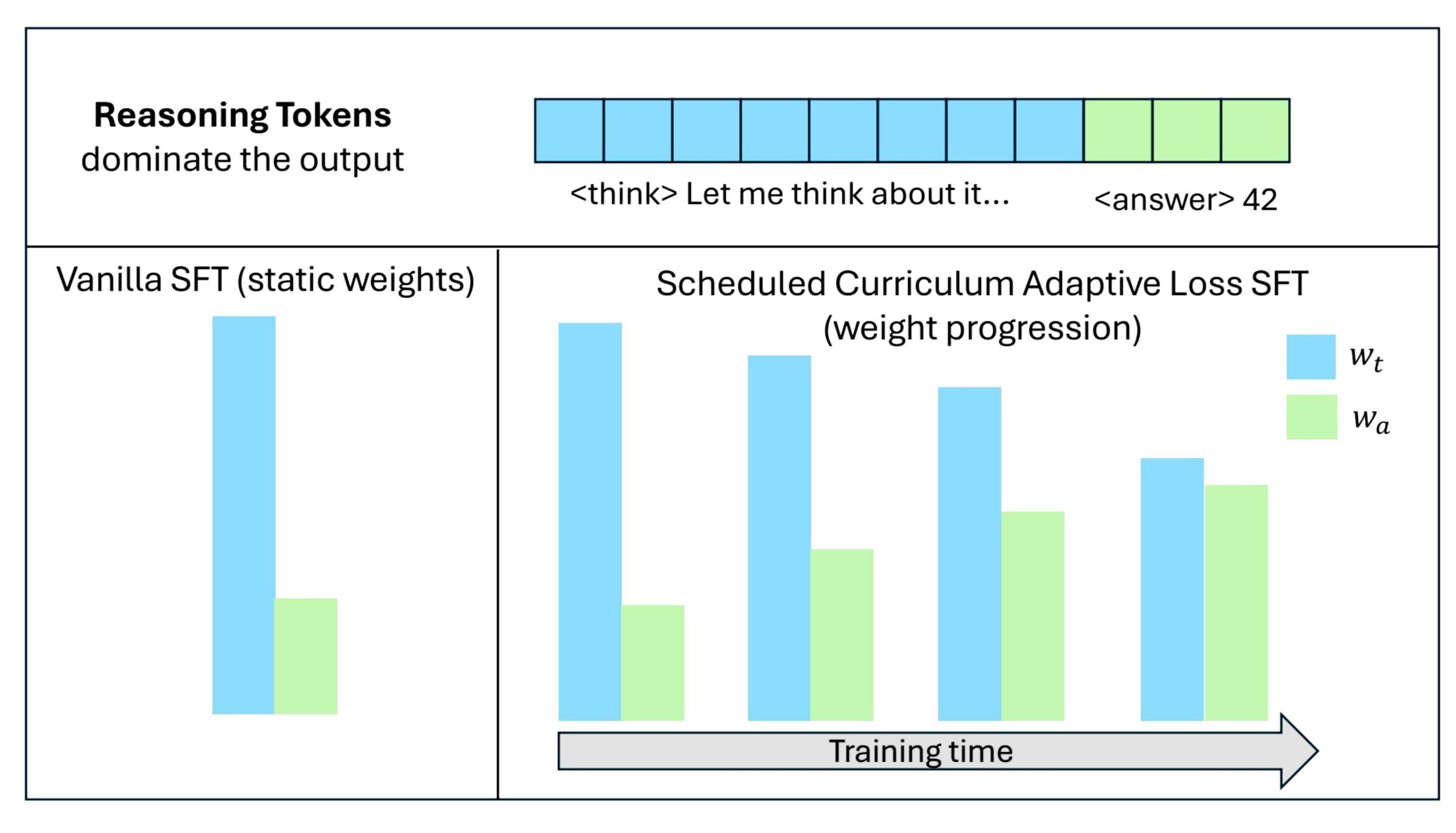}
  \caption{Scheduled Curriculum Adaptive Loss progression over training time vs vanilla SFT. 
  In standard SFT, the \texttt{<think>} portion dominates the output compared to short answer segments (upper panel). SFT applies uniform token weights, making the loss skewed towards the extended reasoning segment (lower left panel). SCALe addresses this token imbalance and introduces a time-dependent weighting schedule that progressively amplifies the influence of answer tokens. As training advances, this shifts the loss toward the final answer, guiding the model to prioritize correct outcomes rather than lengthy reasoning (lower right panel).}
  \label{fig:swl}
  \vspace{-10pt}  
\end{figure}

The development of multimodal reasoning capabilities typically follows a two-phase training pipeline that balances structural supervision with stylistic alignment. \cite{shen2025vlm, luo2025gui, feng2025video}.
In the first phase, supervised fine-tuning (SFT) on reasoning-annotated data establishes a reasoning prior, teaching the model to produce structured and coherent CoTs that connect visual evidence to final answers \cite{huang2025vision}.
The second phase, reinforcement learning (RL) with methods such as Group Relative Policy Optimization (GRPO), refines this behavior by rewarding answer accuracy and preferred response styles \cite{shen2025vlm, guo2025deepseek, plaat2024reasoning}.
Together, these phases promote both the formation and alignment of reasoning skills toward task-specific objectives.

However, despite the importance of SFT, existing approaches overlook the optimization imbalance in reasoning-formatted data.
Standard SFT employs a uniform cross-entropy loss where all tokens contribute equally, regardless of their functional role \cite{ouyang2022training, wei2021finetuned}.
This formulation neglects the asymmetry between the reasoning and answer segments: reasoning sections are typically lengthy and descriptive, while response segments are concise but task-critical \cite{wu2025more, sui2025stop}. 
For example, \cite{huang2025vision} presents a reasoning-oriented dataset, where the average reasoning length is x150 longer than the average response length.
As a result, models become verbose and stylistically fluent but lack factual precision, leading GRPO-initialized checkpoints to favor reasoning style over stable, answer-oriented behavior.

To address these limitations, we propose Scheduled Curriculum Adaptive Loss Supervised Fine-Tuning (SCALe-SFT), an adaptive loss scheduling framework that differentiates between reasoning and answer segments during the supervised training phase.
SCALe-SFT explicitly separates its total loss into reasoning and answer components and assigns controllable weights to each. During training, these weights are dynamically adjusted according to a predefined schedule, initially prioritizing reasoning structure and later emphasizing answer accuracy. 
By decoupling supervision across these segments and progressively scheduling their relative contributions, SCALe-SFT enables the model to gradually build coherent reasoning structures that remain logically consistent and well-aligned with the final answer. 

This concept is illustrated in Figure~\ref{fig:swl}, which shows the model’s predicted tokens. Most belonging to the reasoning stage, with only a small portion corresponding to the final answer. In vanilla SFT, all tokens contribute equally to the loss, resulting in constant weights across reasoning and answer segments. In contrast, our SCALe-SFT dynamically adjusts these weights during training, gradually reducing the influence of reasoning tokens and increasing the emphasis on answer tokens.

We evaluate SCALe-SFT within the Vision-R1 framework \cite{huang2025vision} using its standard two-phase pipeline of reasoning-oriented SFT followed by GRPO. To isolate the effect of our method, we compare it against a vanilla SFT baseline that applies uniform cross-entropy loss across all tokens.
To assess generalization across initialization settings, we examine multiple models, including both instruction-tuned and reasoning-trained variants.
The code will be released upon acceptance.
Our key contributions are summarized as follows:
\begin{itemize}
    \item We identify a core imbalance in reasoning-oriented supervised fine-tuning (SFT), where uniform token-level loss emphasizes lengthy reasoning tokens over its concise answer tokens.
    \item We propose \textbf{Scheduled Curriculum Adaptive Loss Supervised Fine-Tuning (SCALe-SFT)}, a dynamic, segment-aware training framework that progressively balances reasoning and answer supervision through adaptive loss scheduling.
    \item We demonstrate that SCALe-SFT consistently improves accuracy across multiple vision-language models, achieving gains of up to \textbf{3\% after the SFT} phase and \textbf{5\% after the GRPO} phase.
    \item Frequently, SCALe-SFT alone surpasses the performance of standard SFT+GRPO pipelines, offering a lightweight and efficient alternative that allows \textbf{skipping the GRPO phase} when computational resources are limited.
\end{itemize}
\section{Related Work}
\label{sec:related}

\paragraph{Advances in Reasoning in VLMs}
Recent vision-language models (VLMs) have moved beyond single-step perception to reasoning-centric generation, producing intermediate chains that connect visual evidence with textual conclusions \cite{lin2025mind, wang2024measuring}. Modern VLMs typically follow a two-stage training paradigm: supervised fine-tuning on reasoning-annotated data to establish structured reasoning priors, followed by reinforcement learning alignment with task-specific rewards \cite{huang2025vision, shen2025vlm, guo2025deepseek, plaat2024reasoning}.
This combination of structured reasoning and reward-driven optimization has improved performance and interpretability across domains such as medical imaging \cite{lai2025med}, video understanding \cite{feng2025video, zhang2025tinyllava}, and agentic GUI control \cite{zhou2025gui, luo2025gui}, enhancing the safety and reliability of multimodal reasoning systems \cite{zhang2025improve, chen2024measuring}.

\paragraph{Reward-based Reasoning Guidance}
Recent work increasingly focuses on guiding large models’ reasoning during training for greater accuracy and efficiency. Most advances occur at the RL stage, using improved reward functions or optimized algorithms \cite{huang2024post, bu2025beyond, zhao2025bias, dai2025stable, lightman2023let, zhang2025lessons, wen2025reinforcement, su2025crossing}. 
A growing emphasis is placed on shaping the structure and coherence of reasoning rather than merely its surface characteristics. Process-level reward functions evaluate intermediate reasoning steps to encourage coherent thought progression \cite{lightman2023let, zhang2025lessons}, whereas verifier-based rewards (RLVR) replace human preference models with programmatic or verifiable signals to evaluate the correctness of reasoning traces \cite{wen2025reinforcement, su2025crossing}. At the algorithmic level, RL framewords like GRPO has emerged as the dominant paradigm for reasoning-oriented RL. First introduced in DeepSeekMath \cite{shao2024deepseekmath} and later refined in DeepSeek-R1 \cite{guo2025deepseek}, GRPO compares sets of sampled reasoning paths, rewarding those that outperform others in relative terms. This approach improves both exploration efficiency and reward utilization.
Overall, Recent progress centers on reward shaping and RL optimization, while the SFT stage, where models first learn coherent reasoning remains underexplored.


\section{Methods}

We follow the standard Vision-R1 two-phase training pipeline comprising (1) \textbf{supervised fine-tuning (SFT)} and (2) \textbf{reinforcement learning} with GRPO. 
Our work focuses on the SFT stage, where we replace the standard cross-entropy with a scheduled weighted loss: A dynamic weighting strategy that provides a useful inductive bias that stabilizes optimization by balancing the generation of reasoning with its alignment to the final answer. This design enables a VLM to produce coherent reasoning traces and accurate final answers.

\paragraph{Reasoning Dataset}
Recent reasoning-focused VLM research relies on datasets that explicitly couple visual understanding with structured reasoning traces. These datasets extend beyond simple question-answer pairs by including stepwise thought processes annotations.
Each sample contains an image, a textual question, and a reasoning trace delimited by \texttt{<think>...</think>} followed by a final answer enclosed in \texttt{<answer>...</answer>}. 
Notable datasets include Mulberry \cite{yao2024mulberry} and Visual-CoT \cite{shao2024visual}. Most are composed from diverse visual tasks such as captioning, VQA, and chart-understanding datasets, reformatted or synthesized for reasoning-oriented supervision. These datasets are used in the SFT stage to instill reasoning ability in models that lack structured thinking.


\begin{table*}[th]
\centering
\caption{Performance comparison for 3 models, on ScienceQA and IconQA benchmarks.  
Base is the original model without reasoning. SFT denotes vanilla supervised fine-tuning on a reasoning dataset; SCALe denotes our scheduled weighted loss; GRPO and SCALe+GRPO refer to reinforcement stages with initialization from vanilla and weighted-loss SFT, respectively.
The reported metric is an exact match accuracy. We emphasize the best score in bold. SCALe outperforms vanilla SFT across all configurations (up to 3\%) and 5\% after GRPO.}
\label{tab:results}
\vspace{4pt}
\begin{tabular}{lc|c|cc|cc}
\toprule
\textbf{Model} & \textbf{Dataset} & \textbf{Base} & \textbf{SFT} & \textbf{SCALe-SFT} & \textbf{GRPO} & \textbf{SCALe-GRPO} \\

\midrule
\multirow{2}{*}{LLaVA-v1.6-Mistral-7B} 
 & ScienceQA & 72.58 & 80.17 & \textbf{82.6} & 80.86 & \textbf{83.39} \\
 & IconQA    & 37.39 & 67.47 & \textbf{68.25} & 68.21 & \textbf{69.17} \\
\midrule
\multirow{2}{*}{Gemma-3-4b} 
 & ScienceQA & 77.68 & 85.77 & \textbf{85.92} & 85.18 & \textbf{86.37} \\
 & IconQA    & 36.73 & 64.8 & \textbf{65.83} & 64.33 & \textbf{67.8} \\
 \midrule
\multirow{2}{*}{Qwen2.5-VL-3B} 
 & ScienceQA & 80.6 & 85 & \textbf{85.52 }& 85.08 & \textbf{85.82} \\
 & IconQA    & 69.04 & 71.38 & \textbf{71.91} & 74.26 & \textbf{75.58} \\
\bottomrule
\end{tabular}
\end{table*}

\subsection{Scheduled Weighted Loss}
During the SFT stage, we train the model using an autoregressive cross-entropy objective that applies distinct weights to each section of the model's predictions: reasoning and answer tokens.  
In conventional supervised fine-tuning, the loss is defined as a standard cross-entropy (CE) over all predicted tokens, where each token contributes equally:
\begin{equation}
\mathcal{L}_{\text{vanilla}} = \frac{1}{N} \sum_{i=1}^{N} \mathrm{CE}(y_i, \hat{y}_i),
\end{equation}
with $N$ denoting the total number of target tokens.  
While effective for general language modeling, this uniform weighting is suboptimal for reasoning-oriented vision language tasks.  
In such tasks, the model often generates long and verbose reasoning traces enclosed within \texttt{<think>...</think>}, yet evaluation metrics depend primarily on the final short answer segment (\texttt{<answer>...</answer>}).  
As a result, the vanilla CE objective may overemphasize reasoning fluency at the expense of factual correctness, causing the model to ``get lost in thought'', producing extended reasoning that fails to yield accurate answers. In such cases, the wrong answer can be effectively drowned out by the “thinking” loss, with this discrepancy barely reflected in the loss signal.

To address this limitation, we introduce a weighted loss that explicitly separates reasoning and answer supervision. Each output sequence is divided into two semantic segments, the reasoning span enclosed by \texttt{<think>} tags and the answer span enclosed by \texttt{<answer>} tags—allowing each segment to be weighted and optimized independently.

To prevent longer reasoning sequences from dominating the loss, each segment loss is computed in a length-independent manner.  
Let $e_i$ denote the segment (reasoning/answer) for token $i$. The segment loss for a given sample is then:
\begin{equation}
\mathcal{L}_{\text{seg}} = \frac{1}{N_{seg}}\sum_{i: e_i = seg}\mathrm{CE}(y_i, \hat{y}_i)
\end{equation}
where $N_{\text{seg}}$ is the number of tokens of the given segment.  
This formulation computes mean cross-entropy ensuring that longer reasoning traces (e.g., more tokens inside \texttt{<think>...</think>}) do not reduce the influence of shorter examples.

In some cases, reasoning models generate excessively long CoT failing to terminate with a concrete answer. To discourage this behavior, the closing tag \texttt{</think>} is grouped with the subsequent \texttt{<answer>} segment during loss computation.
By treating the closing \texttt{</think>} token as part of the answer segment, the model is further encouraged to end its reasoning deliberately, avoiding unnecessarily verbose or looping thought processes.

To enable a smooth transition between supervision phases, we employ a cosine annealing schedule that gradually interpolates each weight parameter from its initial value to its target value over the course of training. Formally, for a training step $\tau \in [0, T]$, the weight $w(\tau)$ is defined as:
\begin{equation}
\scalebox{0.93}{$
w(\tau) = w^{(\text{end})} + \tfrac{1}{2}\,\bigl(w^{(\text{start})} - w^{(\text{end})}\bigr)\,\Bigl(1 + \cos\!\frac{\pi \tau}{T}\Bigr)$}
\label{eq:cosine_annealing}
\end{equation}
where $w^{(\text{start})}$ and $w^{(\text{end})}$ denote the initial and final target values, respectively, and $T$ is the total number of training steps. 
This formulation ensures that the weighting evolves smoothly, avoiding abrupt transitions that could destabilize optimization. 
At $\tau = 0$, $w(\tau)$ equals its initial value $w^{(\text{start})}$, while at $\tau = T$ it converges exactly to $w^{(\text{end})}$. 

Early in training, higher $w_t$ encourages robust reasoning supervision, while later iterations shift emphasis toward answer accuracy.
Intuitively, we begin training with a higher weight on reasoning tokens to encourage the model to internalize structured, interpretable reasoning patterns early in learning, when generation behavior is unstable. 
As training progresses, the reasoning weight decays toward the answer weight, shifting the objective toward factual correctness once stable reasoning structure has emerged. This behavior is illustrated in Figure~\ref{fig:swl}, where the token importance of the thinking part decreases and the answer part increases over training time. 
This curriculum-like schedule improves both convergence stability and reasoning–answer alignment. The total loss is defined as:
\begin{equation}
\mathcal{L} = w_t \, \mathcal{L}_{\text{think}} + w_a \, \mathcal{L}_{\text{answer}}
\end{equation}
where $\mathcal{L}_{\text{think}}$ and $\mathcal{L}_{\text{answer}}$ denote the mean cross-entropy losses over tokens belonging to the corresponding segments, and $w_t$, $w_a$ are their scheduled weights according to eq.\ref{eq:cosine_annealing}.

\subsection{GRPO Reinforcement Refinement}
Following the SFT stage, an additional GRPO phase can be introduced to further refine reasoning structure and factual correctness, although this stage is not mandatory for the overall training pipeline.
The total reward combines two complementary signals:
\begin{enumerate}
    \item \textbf{Format-structure reward} $r_{\text{tag}}$, which measures the correctness of the reasoning–answer tag placement.  
    Each generated sample is evaluated on four key locations: the opening \texttt{<think>}, the closing \texttt{</think>}, the opening \texttt{<answer>}, and the closing \texttt{</answer>} tag.  
    A partial score of $0.25$ is awarded for each correctly placed tag, yielding:
    \begin{equation}
        r_{\text{tag}} = 0.25 \times n_{\text{correct}},
    \end{equation}
    where $n_{\text{correct}} \in \{0,1,2,3,4\}$ denotes the number of correctly located tags in the output sequence.  
    This structured reward encourages the model to maintain proper reasoning boundaries and to consistently terminate its thought process before generating the final answer.

    \item \textbf{Answer-accuracy reward} $r_{\text{ans}}$, which evaluates an exact match between the generated and reference answers.
\end{enumerate}
The final reward is a weighted combination:
\begin{equation}
r = \lambda_{\text{tag}} \, r_{\text{tag}} + \lambda_{\text{ans}} \, r_{\text{ans}}.
\label{eq:grpo_reward}
\end{equation}
This two-part signal encourages the model to maintain coherent reasoning structure while improving factual correctness.  
Initialization from the weighted-loss checkpoint provides a stable transition into reinforcement optimization, leading to improved reasoning consistency and alignment.

Our two-stage training, scheduled weighted SFT followed by GRPO, jointly shapes how the model reasons and what it answers.  
The length-independent weighted loss promotes interpretable, balanced reasoning during supervised learning, while GRPO fine-tunes reasoning faithfulness and answer accuracy through reinforcement feedback.

\section{Experiments}

\begin{figure*}[t]
\centering
\begin{tcolorbox}[imageFrame, width=\textwidth]
  \begin{minipage}[t]{0.48\textwidth}
  \raisebox{0.3cm}[0pt][0pt]{%
    \begin{tcolorbox}[qaBubble]
      \textbf{Question:} How many motorcycles are there?\\
      \textbf{Answer:} 12
    \end{tcolorbox}
    }
  \end{minipage}\hfill
  \begin{minipage}[t]{0.48\textwidth}
    \centering
    \includegraphics[width=\linewidth]{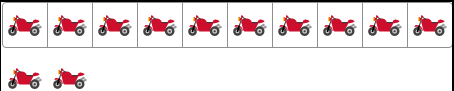}
  \end{minipage}\hfill

\vspace{1mm}
\begin{minipage}[t]{0.48\textwidth}
\begin{tcolorbox}[panelA]
\ttfamily\small
<think>
Okay, let's see. The image illustrates
there are ten red motorcycles arranged
in two rows. The top row has eight
motorcycles, and the bottom row has two.
So, 8 + 2 = 10. The total number should be 10.
</think>\\[2pt]
<answer> Final Answer: 10 </answer>
\end{tcolorbox}
\end{minipage}\hfill
\begin{minipage}[t]{0.48\textwidth}
\begin{tcolorbox}[panelB]
\ttfamily\small
<think>
Okay, let's see. The image displays two
rows of red motorcycles. The top row has ten,
the bottom row has two. So, 10 + 2 = 12.
The total number should be 12.
</think>\\[2pt]
<answer> Final Answer: 12 </answer>
\end{tcolorbox}
\end{minipage}
\end{tcolorbox}
\caption{Reasoning output from Qwen2.5, from HF test set lmms-lab/ICON-QA dataset. Blue (left) box is reasoning from vanilla-SFT. Orange (right) box is SCALe-SFT reasoning. Vanilla SFT counts the number of cars wrong.}
\label{fig:iconqa_ex}
\end{figure*}

\subsection{Experimental Setup}
We compare five configurations: (1) Baseline model before fine tuning (Base), (2) SFT, (3) SCALe-SFT (ours), (4) SFT-GRPO initialized from the SFT checkpoint, and (5) SCALe-GRPO initialized from weighted-loss (SCALe) SFT.
\paragraph{Models.}
We evaluate our proposed weighted loss and GRPO training pipeline on three open-source vision–language models: \textbf{Qwen2.5-VL-3B} \cite{qwen2.5vl2024}, which has undergone prior reasoning-oriented training, and two non-reasoning-pretrained models including \textbf{LLaVA-Next} \cite{liu2024llava} and \textbf{Gemma-3-4B} \cite{gemma32025}.
This setup allows us to examine the adaptability of SCALe-SFT across both reasoning-oriented models and those without existing reasoning priors.


\begin{table*}[hbt]
\centering
\caption{Percentage $\%$ of generations that fail to produce \texttt{<answer>} tags. Lower values indicate better reasoning termination and structural consistency. SCALe and its GRPO substantially reduces invalid outputs compared to standard SFT and GRPO baselines.}
\label{tab:results-validity}
\vspace{4pt}
\begin{tabular}{lc|cc|cc}
\toprule
\textbf{Model} & \textbf{Dataset} & \textbf{SFT} & \textbf{SCALe-SFT} & \textbf{GRPO} & \textbf{SCALe-GRPO} \\

\midrule
\multirow{2}{*}{LLaVA-v1.6-Mistral-7B} 
 & ScienceQA  & 4.76 & \textbf{2.78} & 2.93 & \textbf{1.78} \\
 & IconQA     & 2.54 & \textbf{0.32} & 0.09 & \textbf{0.05} \\
\midrule
\multirow{2}{*}{Gemma-3-4b-it} 
 & ScienceQA  & 2.93 & \textbf{2.08} & 2.68 & \textbf{1.54} \\
 & IconQA     & 2.20 & \textbf{1.68} & 0.05 & \textbf{0.05 } \\
 \midrule
\multirow{2}{*}{Qwen2.5-VL-3B} 
 & ScienceQA  & 2.33 & \textbf{1.54} & 2.58 & \textbf{0.94} \\
 & IconQA     & 0.58 & \textbf{0.53} & 0.01 & \textbf{0} \\
\bottomrule
\end{tabular}
\end{table*}

\paragraph{Datasets.}
We used data from the Vision-R1 reasoning corpus for both SFT and GRPO training.
Vision-R1 \cite{huang2025vision} is a large-scale multimodal reasoning dataset with 200K samples. It is generated via modality bridging pipeline using MLLMs outputs that are in turn fed to DeepSeek-R1 \cite{guo2025deepseek}, yielding structured CoT traces without human annotation.
We evaluate our approach using LMMS-Eval library \cite{lmms_eval2024} on two reasoning-oriented multimodal benchmarks extracted from Vision-R1: \textbf{ScienceQA}~\cite{lu2022scienceqa} and \textbf{IconQA}~\cite{lu2021iconqa}.  
ScienceQA contains image-based and text-only science questions covering natural, social, and language sciences, each paired with multiple-choice answers (A–D) and concise human-written explanations.  
The task requires both visual understanding and scientific reasoning.  
IconQA focuses on diagram and chart comprehension.
Together, these datasets assess both factual reasoning and visual grounding capabilities of vision–language models.


The training set for science related samples from Vision-R1, consisting of ScienceQA and SQA \cite{yao2024mulberry} has $6,593$ samples and for IconQA $15,241$ samples.
The test set consists of $2,017$ samples from ScienceQA. IconQA test set includes three question types: \emph{fill-in-the-blank}, \emph{choose-text}, and \emph{choose-image}.  
Since our models were trained on single-image reasoning, results are reported only for the first two categories and consists of $9,954$ samples. 

\paragraph{Prompt format and answer extraction.}
All experiments used user instructions of the form:  
\emph{``First think about the reasoning process, then provide the answer. The reasoning process and answer are enclosed within \texttt{<think>}...\texttt{</think>} and \texttt{<answer>}...\texttt{</answer>} tags.''}  
During evaluation, we extract the predicted answer from within the \texttt{<answer>} tags.  
Performance is measured by the exact match for both multi-choice and open questions

\paragraph{Training details.} 
In the weighted loss configuration, all models used a cosine scheduling. Reasoning segment weight decreased from $w^{(\text{start})} = 1.0$ to $w^{(\text{end})} =0.5$, while the answer weight remained at $w^{(\text{start})}= w^{(\text{end})} =1.0$.   
Note that even when the weights are fixed, the loss is not the same as vanilla SFT due to its length-independent formulation. For eq. \ref{eq:grpo_reward} we used equal weights for the rewards: $\lambda_{\text{tag}} = \lambda_{\text{ans}} = 1$

\subsection{Results}

Table \ref{tab:results}, shows accuracy comparison for Qwen2.5, LLaVA-Next and Gemma3, on ScienceQA and IconQA benchmarks.
SFT on reasoning-oriented datasets improves performance across all models compared to their pretrained baselines, i.e., the original checkpoints without any additional fine-tuning (the ‘Base’ column in tab~\ref{tab:results}).
On average, SCALe-SFT achieves higher accuracy than vanilla SFT across all three models, confirming that rebalancing the loss between reasoning and answer segments leads to more effective supervision. Notably, in several cases SCALe-SFT even outperformed GRPO initialized from vanilla SFT. Lastly, we show results for GRPO initialized using the improved SCALe-SFT models (marked SCALe-GRPO), which shows an improvement ranging between $0.87\% - 5.4\% $ compared to vanilla GRPO.

Across datasets, the magnitude of improvement varies by model.
On ScienceQA, the proposed SCALe consistently improves reasoning accuracy across all models.
Compared to vanilla SFT, SCALe-SFT achieves an average improvement of up to 3\%.
When combined with GRPO, SCALe-GRPO achieves the highest overall performance, improving upon standard GRPO by an up to 3.12\%, on Llava-Next.
On IconQA, SCALe provides consistent benefits as well.
SCALe-SFT improves over SFT by up to 1.5\%, with gains observed across all three models.
When integrated with GRPO, SCALe-GRPO again yields the best results, improving over GRPO up to 5.4\% on Gemma-3-4B.



These results demonstrate that the proposed SCALe provides a robust, architecture-agnostic mechanism for improving reasoning and answer alignment during SFT, while also stabilizing subsequent GRPO optimization.  
Notably, the strongest results are consistently obtained when GRPO is initialized from SCALe checkpoints, highlighting that reasoning-aware SFT offers an effective foundation for reinforcement refinement.

SCALe formulation implicitly incorporates both reward components later used in GRPO: the higher weights on reasoning and segment boundaries encourage correct tag placement, while the consistent emphasis on the answer segment promotes final-answer accuracy.  
Consequently, the model internalizes structural and factual objectives similar to the reinforcement rewards already during the supervised stage, reducing the marginal gains achieved through subsequent GRPO optimization.  
Therefore, performing only SFT with SCALe offers an effective and efficient alternative to full reinforcement optimization, delivering competitive results at a fraction of the training cost.

\begin{table*}[t]
\centering
\caption{Ablation study on the effect of the scheduled weighting mechanism for Qwen2.5-VL-3B.  
FW (fixed weights) denotes a configuration where all weights remain constant throughout training, effectively disabling the scheduling effect.  
Metrics are Exact Match.}
\label{tab:ablation-schedule}
\vspace{4pt}
\begin{tabular}{lccccc}
\toprule
\textbf{Dataset} & \textbf{SFT} & \textbf{SFT-FW} & \textbf{SCALe-SFT} & \textbf{GRPO-FW} & \textbf{SCALe-GRPO} \\
\midrule
ScienceQA & 85.0 & 85.18 & 85.52  & 84.23 & 85.82 \\
IconQA    &  71.38  & 71.55 & 71.91 & 74.13 & 75.58 \\
\bottomrule
\end{tabular}
\end{table*}

Reasoning models occasionally fail to conclude with a final \texttt{</think>} or even an \texttt{<answer>} tag, resulting in unbounded or incomplete reasoning. We quantify this limitation by measuring the percentage of such invalid generations and show that our SCALe-SFT variant substantially reduces these cases.
Table~\ref{tab:results-validity} reports these rates for all three models tested across ScienceQA and IconQA.
SCALe-SFT reduces invalids in model outputs between 8.62\% to 87.40\% relative reduction compared to vanilla SFT. When looking at the SCALe-GRPO almost has no cases of invalid outputs and the relative reduction from vanilla GRPO is up to 63.5\%.

These results confirm that emphasizing tag tokens within the weighted loss effectively teaches the model to close its reasoning process, yielding structured, well-formed outputs without the need for post-hoc filtering. 
Invalid outputs are not only undesirable for evaluation but also costly in real-world systems, where unclosed or endlessly generating responses often require rerunning inference or manual recovery.  
By substantially reducing invalid generations, SCALe improves both the reliability and sustainability of model deployment in practical reasoning pipelines.

\subsection{Ablation Study}

To isolate the contribution of the scheduling mechanism in our weighted loss, we conduct an ablation on the Qwen2.5-VL-3B model using both the ScienceQA and IconQA benchmarks.  
In this experiment, we compare our SCALe-SFT configuration, where the reasoning weight decays over training, with a control variant in which the schedule is effectively neutral, starting and ending at identical weights of $1.0$.  
This variant is denoted as FW (fixed weights) and represents a case where scheduling has no influence on token weighting.  
We evaluate both the SFT and GRPO stages under these two setups.

\begin{figure*}[t]
\centering

\begin{tcolorbox}[imageFrame, width=\textwidth]
    \begin{minipage}[t]{0.48\textwidth}
  \raisebox{1cm}[0pt][0pt]{%
    \begin{tcolorbox}[qaBubble]
      \textbf{Question:} Is the following statement about our solar system true or false?\\
      50\% of the planets are made mainly of gas.\\[3pt]
      \textbf{Choices:} \texttt{['true', 'false']}\\
      \textbf{Answer:} \textbf{B}
    \end{tcolorbox}
    }
  \end{minipage}\hfill
  \begin{minipage}[t]{0.48\textwidth}
    \centering
    \includegraphics[width=0.7\linewidth]{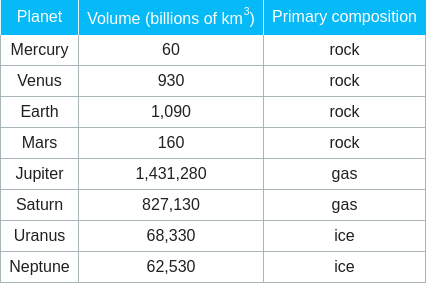}
  \end{minipage}
  \vspace{-1mm}

  \begin{minipage}[t]{0.48\textwidth}
    \begin{tcolorbox}[panelA]
      \ttfamily\small
      <think>
      Okay, let me try to figure this out. The question is asking whether 50\% of the planets in our solar system are made mainly of gas. From the image provided, the planets listed are Mercury, Venus, Earth, Mars, Jupiter, Saturn, Uranus, and Neptune. The primary compositions are rock for Mercury, Venus, Earth, and Mars; gas for Jupiter and Saturn; ice for Uranus and Neptune. So, out of the eight planets mentioned, only two (Jupiter and Saturn) are gas giants. That makes four planets that are not gas. So, if 2 out of 8 are gas, then 4/8 = 50\%. Therefore, the statement should be true.
      </think>\\[2pt]
      <answer> Final Answer: \textbf{A}. </answer>
    \end{tcolorbox}
  \end{minipage}\hfill
  \begin{minipage}[t]{0.48\textwidth}
    \begin{tcolorbox}[panelB]
      \ttfamily\small
      <think>
      Okay, let's see. The question is asking whether 50\% of the planets in our solar system are made mainly of gas. From the image provided, there are eight planets listed. The primary compositions are rock for Mercury, Venus, Earth, Mars; gas for Jupiter and Saturn; ice for Uranus and Neptune. So that makes two planets made of gas (Jupiter and Saturn) and two made of ice (Uranus and Neptune). That's a total of eight planets. If 50\% of them were gas, then half would be 4 out of 8. But here, only 2 are gas. So the statement is false.
      </think>\\[2pt]
      <answer> Final Answer: \textbf{B}. </answer>
    \end{tcolorbox}
  \end{minipage}

\end{tcolorbox}

\caption{Reasoning output from Qwen2.5, from HF test set lmms-lab/ScienceQA-IMG dataset. Blue bottom right box is reasoning from vanilla-SFT. Orange bottom left is SCALe-SFT reasoning. Vanilla SFT correctly counts two gas giants among eight planets yet concludes that 50\% are gas, revealing a logical inconsistency between reasoning and answer, while Scheduled Weighted SFT maintains logical consistency and yields the correct conclusion.}
\label{fig:scienceqa_ex}
\end{figure*}

Table~\ref{tab:ablation-schedule} shows that incorporating a scheduled decay of the reasoning weight provides consistent gains across both datasets.  
The SFT-FW variant, is still better than vanilla SFT, but the full SCALe-SFT has a relative improved average of 0.45\%. 
The absence of scheduling also limits the benefit of GRPO.  
While GRPO-FW yields an improvement over SFT, the gain remains smaller than with full SCALe-GRPO, which has an average improvment of 1.92\%.  

Without a progressive decay of the reasoning weight, the SFT stage fails to establish a clear balance between reasoning and answer supervision, resulting in reasoning traces that are noisier or less distinctly segmented.  
Consequently, during GRPO training—where rewards are concentrated around tag placement and answer accuracy—the model receives high-variance reinforcement signals that can over-correct or truncate its reasoning, leading to marginal degradation.  
In contrast, the scheduled SCALe pretraining provides a smoother structural prior that aligns naturally with GRPO rewards, enabling stable optimization and consistent gains.
  
These results confirm that the gradual scheduling of reasoning weight acts as a lightweight curriculum mechanism, guiding the model to first focus on structured reasoning and later refine answer precision.

\subsection{Qualitative Examples}

To better illustrate the effect of our weighted loss training, we present qualitative examples of visual reasoning from Qwen2.5 outputs on tested datasets in Figs.~\ref{fig:scienceqa_ex}-\ref{fig:iconqa_ex}.  
Each example shows the input image, the question, choices (if present), ground-truth answer, and the model’s generated reasoning trace and final answer.  
We compare generations from vanilla SFT (light blue box in each image example) and our proposed SCALe-SFT setup (light orange box). 
Additional qualitative results are provided in Figs.~\ref{fig:gemma3-scienceqa_ex}-\ref{fig:llave-scienceqa_ex} of the supplementary material.
Typical failure cases include non-convergent reasoning loops, semantically sound reasoning that concludes with an incorrect answer, and weaker alignment between the reasoning trace and the final prediction.

As can be seen from figure ~\ref{fig:iconqa_ex}, uniform SFT loss trains the model to produce plausible reasoning traces but not to verify visual accuracy, so it may “hallucinate” object counts. Scheduled Weighted SFT progressively ties reasoning fidelity to answer correctness, encouraging perceptually grounded reasoning: the model actually counts the ten motorcycles before concluding “10 + 2 = 12.”
In figure ~\ref{fig:scienceqa_ex}, vanilla SFT model correctly identifies that only two of the eight planets are gas giants but then misinterprets this as 50\%, producing a self-contradictory wrong conclusion despite a fluent reasoning trace. It prioritize “sounding like reasoning” over reasoning accuracy, since both contribute equally to the loss.

The SCALe-trained model produces more structured reasoning enclosed by \texttt{<think>} and \texttt{<answer>} tags, often arriving at correct answers through concise, interpretable reasoning steps.  
In contrast, the vanilla SFT model tends to generate long or disorganized reasoning, occasionally leading to incorrect conclusions.
Scheduled Weighted SFT doesn’t change the supervision data, it changes how the model learns from it. As training progresses, the loss gradually shifts emphasis toward answer-related tokens, teaching the model to align its reasoning with correct final outcomes.



\section{Conclusion}

We introduced a \textbf{SCALe-SFT} – Scheduled Curriculum Adaptive Loss SFT for vision–language reasoning, a simple modification to supervised fine-tuning that applies dynamic, length-independent weighting over reasoning and answer spans.

The proposed SCALe-SFT encourages concise and well-structured reasoning, leading to more accurate final answers. It serves both as a lightweight alternative to reinforcement learning and as a stronger initialization for subsequent GRPO optimization using tag-location and answer-accuracy rewards, resulting in improved reasoning alignment.

Experiments show consistent gains over vanilla SFT and GRPO alone. 
Termination analysis shows SCALe-SFT sharply reduces invalid generations (unclosed \texttt{<answer>}), thereby reducing the need for repeated inference to recover valid outputs.
Ablation study indicates that replacing the cosine schedule with fixed weights reduced performance gains, confirming that dynamic weighting is better for effective reasoning supervision.

Future work will extend the proposed framework to a broader range of and modalities and multi-image cases.
Another research direction will be to explore an automatic online weight tuning.
We also plan to apply our method to language-only reasoning datasets.  
Since SCALe-SFT operates purely on token-level supervision, it can be directly applied to textual reasoning tasks without modification, providing a unified approach to structured reasoning across visual and linguistic domains.  
Such extensions will help assess whether the benefits observed here, improved reasoning structure, reduced invalid terminations, and enhanced answer alignment, generalize beyond visual question answering to broader reasoning scenarios.

{
    \small
    \bibliographystyle{unsrtnat} 
    \bibliography{main}
}


\clearpage
\appendix
\setcounter{page}{1}
\setcounter{section}{0}
\maketitlesupplementary


\section{Overview}
\label{sec:overview}
This supplementary document provides additional results and implementation details
that complement the main paper. Specifically, we include:
\begin{itemize}
    \item Qualitative comparisons between SFT, SCALe, reasoning traces illustrating improved factual grounding for Gemma-3-4b and Llava-next.
    \item Implementation Details for SFT and GRPO.

\end{itemize}

\section{Additional Qualitative Results}
\label{sec:qualitative}
\begin{figure*}[t]
\centering

\begin{tcolorbox}[imageFrame, width=\textwidth]
    \begin{minipage}[t]{0.48\textwidth}
  \raisebox{1cm}[0pt][0pt]{%
    \begin{tcolorbox}[qaBubble]
      \textbf{Question:} Which animal is also adapted to use its neck to appear large and scary to a predator?\\[3pt]
      \textbf{Choices:} ['bearded dragon', 'green anole']\\
      \textbf{Answer:} \textbf{A}
    \end{tcolorbox}
    }
  \end{minipage}\hfill
  \begin{minipage}[t]{0.48\textwidth}
    \centering
    \includegraphics[width=0.6\linewidth]{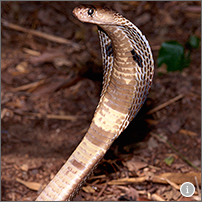}
  \end{minipage}
  \vspace{-1mm}

  \begin{minipage}[t]{0.48\textwidth}
    \begin{tcolorbox}[panelA]
      \ttfamily\small
      <think>
      Okay, let's see. The question is asking which animal, besides the spectacled cobra, also uses its neck to appear large and scary to a predator. The options are A. bearded dragon and B. green anole. From the context given, the spectacled cobra uses its neck to appear large and scary. The answer should be another animal that does the same. I remember that bearded dragons and green anoles are both lizards. Lizards, in general, can puff up their bodies, including their necks, to look bigger. For example, bearded dragons sometimes puff up their necks to appear larger, which might scare off predators. Green anoles are known for their ability to flatten their bodies and raise their necks, which can make them look bigger. So both bearded dragons and green anoles are likely to use their necks in a similar way. The question is asking which animal is also adapted to do this. Since both options are lizards that can puff up their necks, the answer should be either A or B. The image also shows a spectacled cobra, so the answer is another animal that does the same. The options are A and B, so the answer is either A or B. Since the question is asking which animal is also adapted, both A and B are options. However, the question is phrased as "which animal is also adapted," so the answer would be either A or B. But the options are A and B. The answer should be either A or B. Since the question is asking which animal is also adapted, both A and B are options. But the question is phrased as "which animal...
      
    \end{tcolorbox}
  \end{minipage}\hfill
  \begin{minipage}[t]{0.48\textwidth}
    \begin{tcolorbox}[panelB]
      \ttfamily\small
      <think>
      Okay, let's see. The question is asking which animal, besides the spectacled cobra, uses its neck to appear large and scary to a predator. The options are A. bearded dragon and B. green anole. From the image given, the spectacled cobra uses its neck to appear large and scary. The question is asking which other animal does this. The bearded dragon and green anole are both reptiles, but I need to recall if either of them uses their neck in a similar way. I remember that bearded dragons, especially males during mating displays, can puff up their necks to appear larger. They also have crests that they can raise or lower. Green anoles, on the other hand, are more about camouflage and quick movements. They don't have prominent neck displays like bearded dragons. So the bearded dragon is more likely to use its neck for a display, while the green anole doesn't. Therefore, the answer should be A. bearded dragon.\\
      </think>\\[2pt]
      <answer> Final Answer: \textbf{A}. </answer>
    \end{tcolorbox}
  \end{minipage}

\end{tcolorbox}

\caption{Reasoning output from Gemma3, from HF test set lmms-lab/ScienceQA-IMG dataset. Blue bottom right box is reasoning from vanilla-SFT. Orange bottom left is SCALe-SFT reasoning. Vanilla SFT goes into an endless loop and cannot reach a final answer (shortened with ...) while SCALe-SFT has consice reasoning steps and accurate final answer.}
\label{fig:gemma3-scienceqa_ex}
\end{figure*}

\begin{figure*}[t]
\centering

\begin{tcolorbox}[imageFrame, width=\textwidth]
    \begin{minipage}[t]{0.48\textwidth}
  \raisebox{1cm}[0pt][0pt]{%
    \begin{tcolorbox}[qaBubble]
      \textbf{Question:} Which country is highlighted?\\[3pt]
      \textbf{Choices:} \texttt{['Saint Vincent and the Grenadines', 'Grenada', 'the Dominican Republic', 'Trinidad and Tobago']}\\
      \textbf{Answer:} \textbf{B}
    \end{tcolorbox}
    }
  \end{minipage}\hfill
  \begin{minipage}[t]{0.48\textwidth}
    \centering
    \includegraphics[width=0.7\linewidth]{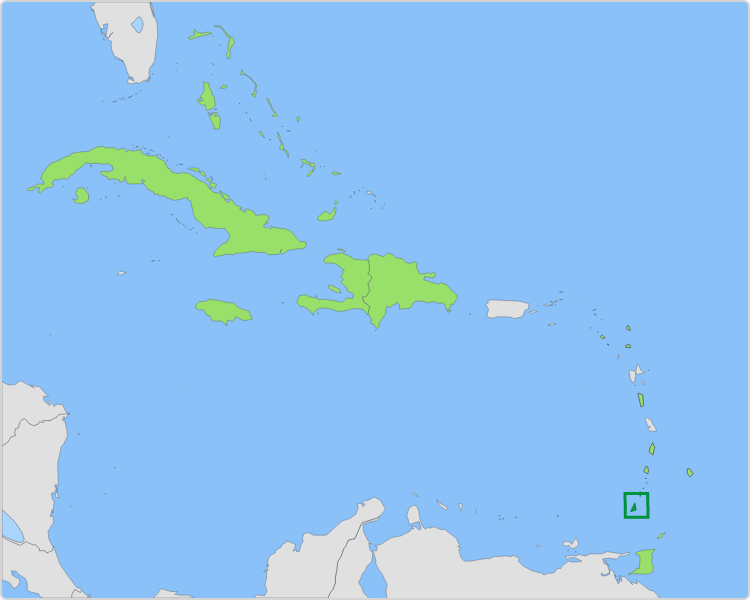}
  \end{minipage}
  \vspace{-1mm}

  \begin{minipage}[t]{0.48\textwidth}
    \begin{tcolorbox}[panelA]
      \ttfamily\small
      <think>
      Okay, let's see. The question is asking which country is highlighted on the map of the Caribbean. The image illustrates depicts the highlighted area is in the southern part of the Lesser Antilles, near the northeastern coast of South America. The options are Saint Vincent and the Grenadines, Grenada, the Dominican Republic, and Trinidad and Tobago. First, I need to recall the geography of the Caribbean. The Lesser Antilles are a group of islands in the Caribbean Sea. The southern part of the Lesser Antilles includes countries like Grenada, Saint Vincent and the Grenadines, and Trinidad and Tobago. The Dominican Republic is on the island of Hispaniola, which is more to the north, part of the Greater Antilles. The image illustrates displays the highlighted area is near the northeastern coast of South America. Trinidad and Tobago are located just off the coast of Venezuela, which is in South America. So if the highlighted area is near South America, that points to Trinidad and Tobago. The other options are further north in the Lesser Antilles. Grenada is also in the southern part but a bit more to the north compared to Trinidad and Tobago. Saint Vincent and the Grenadines are further north as well. So the answer should be D. Trinidad and Tobago.
      </think>\\[2pt]
      <answer> Final Answer: \textbf{D}. </answer>
    \end{tcolorbox}
  \end{minipage}\hfill
  \begin{minipage}[t]{0.48\textwidth}
    \begin{tcolorbox}[panelB]
      \ttfamily\small
      <think>
      Okay, let's see. The question is asking which country is highlighted on the map of the Caribbean. The image illustrates depicts the highlighted country is in the southern Caribbean, near the northeastern coast of South America. It's south of Saint Vincent and the Grenadines and north of Trinidad and Tobago. The options are A to D. First, I need to recall the geography of the Caribbean. The southern Caribbean includes countries like Venezuela, Colombia, and the islands of the Lesser Antilles. The Grenadines are part of Saint Vincent and the Grenadines, which is one of the Windward Islands. So if the highlighted country is south of Saint Vincent and the Grenadines, that would place it in the southern part of the Lesser Antilles. Trinidad and Tobago are further south, near Venezuela. The Dominican Republic is on Hispaniola, which is more to the north, near Haiti. So the country between Saint Vincent and the Grenadines and Trinidad and Tobago would be Grenada. The answer should be B.
      </think>\\[2pt]
      <answer> Final Answer: \textbf{B}. </answer>
    \end{tcolorbox}
  \end{minipage}

\end{tcolorbox}

\caption{Reasoning output from Llava-next, from HF test set lmms-lab/ScienceQA-IMG dataset. Blue bottom right box is reasoning from vanilla-SFT. Orange bottom left is SCALe-SFT reasoning. In vanilla SFT the token imbalance can lead to well structured reasoning traces that fail to align with the final answer supervision-     the model “thinks correctly” but concludes incorrectly.}
\label{fig:llave-scienceqa_ex}
\end{figure*}

Figures~\ref{fig:gemma3-scienceqa_ex} - \ref{fig:llave-scienceqa_ex} present additional examples comparing the reasoning structures produced by the different STF variations from the models compared in the main paper. All examples are taken from HF test set lmms-lab/ScienceQA-IMG dataset. Blue bottom right box is reasoning trace from vanilla-SFT and orange bottom left shows SCALe-SFT reasoning.

Our SCALe model demonstrates more coherent intermediate reasoning steps, encourages the model to complete its reasoning trace coherently and converge toward the correct conclusion, resulting in more reliable \texttt{<think>} endings and answer transitions compared to standard SFT.
This can be seen by fig~\ref{fig:gemma3-scienceqa_ex}, where vanilla SFT goes into a reasoning loop, where it endlessly debates about the answer and fails to reach a final conclusion.
In contrast, fig~\ref{fig:llave-scienceqa_ex} shows the vanilla SFT often can produce coherent reasoning trace but incorrect final answer, 
reflecting the token imbalance that biases optimization toward verbose reasoning segments. 


\section{Implementation Details}
\label{sec:impl}
\paragraph{Training.}
All models were trained using \texttt{Deepspeed-ZeRO2} on 8$\times$H100 GPUs.  
SFT training ran for one epoch over the respective Vision-R1 subsets.   
We used mixed precision (\texttt{bfloat16}) and gradient checkpointing.
All models were fine-tuned using LoRA adapters with a learning rate of $1\!\times\!10^{-4}$ and a maximum sequence length of $8192$ tokens.  
For the RL stage, we employed GRPO with a maximum generation length of $1024$ tokens, generating $8$ completions per prompt using sampling with temperature $1.0$ (top-$p=1.0$, no top-$k$ filtering). Inference was accelerated using vLLM in colocate mode.

\paragraph{Evaluation.}
We evaluated all models using the \texttt{lmms-eval} framework.
To accommodate reasoning-based responses, we extended \texttt{lmms-eval} to parse the final answer from within the \texttt{<answer>} tags.
All metrics were calculated exclusively on the extracted final answers.
We used greedy decoding (temperature $0$) for all evaluations to ensure deterministic and reproducible results.

\clearpage


\end{document}